\documentclass[10pt, a4paper]{article}

\usepackage[final]{lrec2026} 
\usepackage{todonotes}
\usepackage{graphicx}
\usepackage{booktabs}
\usepackage[most]{tcolorbox}
\tcbset{
  promptbox/.style={
    colback=gray!10,
    colframe=black,
    boxrule=0.5pt,
    arc=4pt,
    left=6pt,
    right=6pt,
    top=6pt,
    bottom=6pt
  }
}

\usepackage[svgnames]{xcolor}

\definecolor{mygreen}{HTML}{D5E8D4}
\definecolor{myorange}{HTML}{FFE6CC}
\definecolor{myred}{HTML}{F8CECC}

\usepackage{stfloats} 

\usepackage{textcomp} 

\title{A Catalog of Basque Dialectal Resources: \\ Online Collections and Standard-to-Dialectal Adaptations}

\name{Jaione Bengoetxea, Itziar Gonzalez-Dios, Rodrigo Agerri} 

\address{HiTZ Center - Ixa, University of the Basque Country EHU \\
         \{jaione.bengoetxea,itziar.gonzalezd,rodrigo.agerri\}@ehu.eus\\}

\abstract{
Recent research on dialectal NLP has identified data scarcity as a primary limitation. To address this limitation, this paper presents a catalog of contemporary Basque dialectal data and resources, offering a systematic and comprehensive compilation of the dialectal data currently available in Basque. Two types of data sources have been distinguished: online data originally written in some dialect, and standard-to-dialect adapted data. The former includes all dialectal data that can be found online, such as news and radio sites, informal tweets, as well as online resources such as dictionaries, atlases, grammar rules, or videos. The latter consists of data that has been adapted from the standard variety to dialectal varieties, either manually or automatically. 
Regarding the manual adaptation, the test split of the XNLI Natural Language Inference dataset was manually adapted into three Basque dialects: Western, Central, and Navarrese-Lapurdian, yielding a high-quality parallel gold standard evaluation dataset. With respect to the automatic dialectal adaptation, the automatically adapted physical commonsense dataset (BasPhyCo\textsubscript{west}) underwent additional manual evaluation by native speakers to assess its quality and determine whether it could serve as a viable substitute for full manual adaptation (i.e., silver data creation).
 \\ \newline \Keywords{Basque, dialects, low-resource, data-collection} }
 
\begin{document}

\maketitleabstract

\section{Introduction}

Dialectal variation is a core feature of all natural languages. However, up until now, Natural Language Processing (NLP) research has almost exclusively focused on tailoring data and resources for the standard forms of each language. 

In recent years, this trend has slowly started to shift, with some studies increasingly focusing on dialects. However, the range of tasks and languages addressed remains fairly limited. Several surveys review recent developments and research directions in dialectal NLP, such as \citet{joshi2025survey} and \citet{Zampieri_Nakov_Scherrer_2020}.

Therefore, a major limitation identified by these works is the scarcity of data and resources for non-standard varieties. This presents a significant challenge, since recent advances in state-of-the-art NLP have reinforced the importance of data quantity in developing high-performing language technology tools, especially in low-resource scenarios  \citep{artetxe-etal-2022-corpus}. Given the importance of data quantity, the field of dialectal NLP could be regarded as a low-resource research scenario. 


The lack of modern dialectal data is especially pronounced for Basque NLP technologies. The majority of resources have been developed with a high focus on Standard Basque, such as spell-checkers \citep{agirre-etal-1992-xuxen}, Neural Machine Translators\footnote{\href{https://elia.eus/traductor}{Elia} or \href{https://www.euskadi.eus/itzuli/}{Itzuli}.} or, more recently, text representations \cite{agerri-etal-2020-give}, and instruction fine-tuned Large Language Models (LLMs) trained for Basque \citep{etxaniz-etal-2024-latxa, sainz-etal-2025-instructing}.


In this context, this paper aims to present a thorough compilation of Basque dialectal data and resources in order to facilitate potential future work on Basque dialects. We hope this will be useful not only in the field of NLP, but also in other related areas such as sociolinguistics or variationist linguistics. 

This work groups Basque dialectal data and resources into two main categories, presented in Section \ref{sec:online-data} and Section \ref{sec:adaptations}, respectively:

\begin{figure}[!t]
  \centering
  \includegraphics[width=0.7\linewidth]{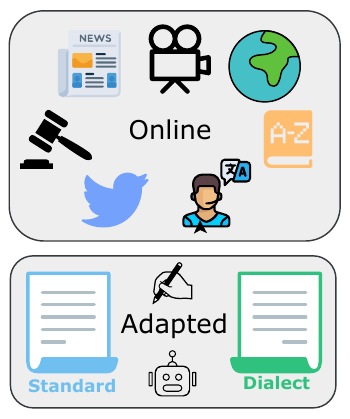}
  \caption{The Basque dialectal catalog consists of two different sources: \textbf{online} dialectal data and standard-to-dialect \textbf{adapted} data, either manually or automatically.}
  \label{fig:catalog}
\end{figure}

\begin{itemize}
    \item \textbf{Online dialectal data and resources.} This category includes online sites such as local news and radio stations, tweets, linguistic atlases, dictionaries, and videos. 
    \item \textbf{Standard-to-dialect adapted data.} This includes standard data adapted into dialects, either manually or automatically. For the former, we present \textbf{Parallel XNLIvar}, an expansion of XNLIvar \citep{bengoetxea-etal-2025-lost} that includes the manual adaptation of the 5000-instance test set for the task of Natural Language Inference (NLI). The latter describes \textbf{BasPhyCo}, a physical commonsense dataset automatically adapted into dialectal Basque using Large Language Models (LLMS) \citep{bengoetxea2026physical}. A novel manual evaluation has been conducted by native dialect speakers on the automatically obtained data.
\end{itemize}

These resources and adapted datasets are collected and publicly available\footnote{\url{https://anonymous.4open.science/r/Catalog-of-Basque-Dialects-D87A}}.

\section{Related Work}
\label{sec:related-work}

\paragraph{Dialectal Data Collection}

Many recent studies have focused their research on the systematic collection of dialectal data as a strategy to address the problem of data scarcity. For instance, \citet{sun-etal-2025-dia} introduced a gamified dialect data collection system, where native speakers could access an interface and either rewrite sentences into their dialect or match dialectal sentences to their geographical location. Their work has been found to efficiently increase user engagement in data collection processes. 

However, given the growing demand to acquire large volumes of data, manual collection is frequently impractical. Therefore, several works have resorted to silver data creation, such as Multi-VALUE \citep{ziems-etal-2023-multi}, a rule-based translation system for 50 different English dialects and 189 linguistic features, which performs as a successful data augmentation method. 

Although the critical role of dialectal data is widely recognized, no study currently provides a comprehensive dialectal database.
There are, however, some benchmark works that do not explicitly focus on data, but provide an invaluable source of dialectal data. For instance, \citet{FAISAL2024DIALECTBENCHAN} presented an extensive dialectal benchmark, evaluating 10 different tasks in 281 varieties. Additionally, \citet{alam-etal-2024-codet} focused their dialectal benchmark on Machine Translation (MT), as they evaluated several varieties from 12 different languages. 

Both of these benchmarks included Basyque \citep{uria2012hizkeren}, a dataset of Northern Basque dialects used to evaluate MT. No other task was evaluated in Basque in these works, as there was no other dialectal dataset for modern Basque dialects at the time of these benchmarks. 

\paragraph{Basque Dialects} 

The classification of Basque dialects has been up for debate for many years. 
\citet{bonaparte1869verbe} proposed a classification of eight dialects and 25 subdialects, a distribution that highlights the remarkable degree of variation within Basque, despite its comparatively small geographic area.


This classification has been considered canonical until a recent Basque dialectology work by \citet{zuazu2008}, which established an extensive and comprehensive categorization of five Basque dialects, and provided a broad archive of the most representative features of each variation.

This paper will adopt the dialectal classification proposed by \citet{zuazu2008}, distinguishing the following Basque dialects: Western, Central, Navarrese, Navarrese-Lapurdian and Zuberoan. For the purposes of this work, the Navarrese-Lapurdian and Zuberoan dialects will additionally be referred to as Northern dialects. A map illustrating this classification by \citet{zuazu2008} is provided in the Appendix. 

\paragraph{Basque Standardization}

While extensive dialectal diversity may constitute a source of linguistic richness, it can also increase the risk of language endangerment. This concern led many Basque linguists to push for the necessity of a standard variation \citep{elkartea2010hizkuntzaren}. 

The process for the creation of a standard variation was a lengthy one. Many meetings were held by contemporary linguists, until Koldo Mitxelena's proposal in the 60s. He suggested using the Central dialects as the foundation of the Standard, mainly due to its practicality: it was the dialect with the most literary prestige to date, as well as being understandable by all Basque speakers. This proposal was discussed, modified, and accepted in the Congress of Arantzazu in 1968 \citep{elkartea2010hizkuntzaren}. 




The emergence of the standard variety was followed by growing movements to preserve and revitalize dialects. For instance, the Standard Western variety, which was developed to formalize and thus promote its use in written form as well as other registers and use cases \citep{ikastegia2001bizkai}. 


Overall, Basque standardization and dialect preservation have developed side by side, both playing an important role in shaping the language today. This work aims to contribute to this movement by providing a collection of current dialectal resources.

\begin{table*}
    \centering

    \begin{tabular}{@{}lrl rl r@{}}
    
    \toprule
        \textbf{Source} & \textbf{Type} & \textbf{Dialect} & \textbf{Register} & \textbf{Modality} & \textbf{License} \\ \midrule 
         \href{https://bizkaiairratia.eus/}{Bizkaia Irratia} & Text \& Audio & Western & Formal & News \& Radio & cc-by-sa \\
         \href{https://bizkaie.biz/}{Bizkaie!} & Text \& Audio & Western & Formal & News \& Radio & CC-BY-SA \\
         \href{https://xiberokobotza.org/}{Xiberoko Botza} & Text \& Audio & Zuberoan & Formal & News \& Radio & CC-BY-SA \\
         \href{https://irulegikoirratia.eus/}{Irulegiko Irratia} & Text \& Audio & Nav-Lap & Formal & News \& Radio & CC-BY-SA \\ 
         \href{https://gureirratia.eus/}{Gure Irratia} & Text \& Audio & Nav-Lap & Formal & News \& Radio & CC-BY-SA\\
         \href{https://jjggbizkaia.eus/eu/hasierea}{General Assemblies} & Text & Western & Specialized & Minutes & N/A \\ 
         \href{https://huggingface.co/datasets/HiTZ/BERnaT-Diverse}{BSM} & Text & Mixed & Informal & Tweets &  CC-BY-SA \\
         \midrule \midrule
         \href{https://www.euskaltzaindia.eus/index.php?option=com_content&view=article&id=565&Itemid=466&lang=eu}{Linguistic Atlas} & Resource & Mixed & - & Atlas & N/A \\ 
         \href{https://www.euskaltzaindia.eus/index.php?option=com_hiztegianbilatu&task=hasiera&Itemid=1693&lang=eu}{Euskaltzaindia} & Resource  & Mixed & - & Dict. & N/A \\
         \href{https://hiztegia.labayru.eus/?locale=es}{LabayruHiztegia} & Resource & Western &  - & Dict. & N/A \\
         \href{https://gramatika.labayru.eus/}{LabayruGramatika} & Resource & Western & - & Grammar & N/A \\ 
         \href{https://ixa2.si.ehu.eus/atlas2/help.php?lang=eu}{Basyque} & Resource & Northern & - & Grammar & CC-BY-SA \\
         \midrule \midrule
         \href{https://ahotsak.eus/corpusa/}{Ahotsak} & Video & Mixed & - & Speech & CC-BY-SA \\ 
         \href{https://www.mintzoak.eus/eu/}{Mintzoak} & Video & Northern & - & Speech & CC-BY-NC-SA\\
         \href{http://euskalkiak.eus/ikus_entzunezkoak.php}{Euskalkiak.eus} & Video & Mixed & - & Speech
         & CC-BY-SA \\
         \bottomrule
    \end{tabular}
    \caption{Summary of online Basque dialectal resources. N/A = Not Available.}
    \label{tab:online-sources}
\end{table*}

\section{Online Dialectal Data}
\label{sec:online-data}

In this section, we introduce the dialectal data available within the Basque online community, categorized according to their modality. All sources and their characteristics are summarized in Table \ref{tab:online-sources}. 

\subsection{News and Radio Sites}
\label{sec:news-radio}

Several Basque news sites write articles in their local dialectal variety. For instance, news sites that have been written in the Western dialect include \href{https://bizkaie.biz/}{\textbf{Bizkaia Irratia}}
and \href{https://bizkaie.biz/}{\textbf{Bizkaie!}}.
Inside the news domain, these texts are written in a formal, not technical register.  

These sites provide up-to-date news articles in the Standard Western dialect. 
This variety was first formalized in \citet{ikastegia2001bizkai}, where orthography, morphology (including verb declination), and syntax issues were established. More recently, the Labayru foundation has digitalized this grammar and provided Standard Western Basque information on their website (more information on Section \ref{sec:grammar}).

Apart from the Western dialect, several online news sites are also available in other varieties, including the Northern dialects. In fact, the \textbf{Euskal Irratiak} association consists of several independent radios that collect news from different locations, in their local variation of Basque. These include \href{https://xiberokobotza.org/}{\textbf{Xiberoko Botza}}, \href{https://irulegikoirratia.eus/}{\textbf{Irulegiko Irratia}} and \href{https://gureirratia.eus/}{\textbf{Gure Irratia}}.


\subsection{Legal Documents}
\label{sec:assemblies}

The website of the \href{https://jjggbizkaia.eus/eu/hasierea}{\textbf{Biscayan General Assemblies}}
contains the minutes of the highest organizational body representing the citizens of Biscay, i.e., the Biscayan parliament. This organization exercises regulatory power in the region and approves the budgets for the historical territory.

Some of the legal documents on their website are written in the Western dialect, providing dialectal texts not only in a formal register, but also in the specialized legal domain. As is the case for the news site texts, these technical documents are written in the Standard Western dialect. 

\subsection{Basque Social Media Corpus}
\label{sec:social-media}

Some previous work has been done on the comparison of different registers in Basque, where informal speech often includes strong dialectal variations. These works include \citet{fernandezdealanda-2019} and \citet{FernandezdeLanda2021}, where they work on real-world data collected from Twitter, which includes dialectal, slang, informal, and code-switched data. The \href{https://huggingface.co/datasets/HiTZ/BERnaT-Diverse}{\textbf{Basque Social Media (BSM)}}
corpus consists of approximately 11 million posts produced by more than 13,000 Basque-speaking users, with a total of around 188 million words.

\subsection{Linguistic Atlas}
\label{sec:linguistic-atlas}

Euskaltzaindia is the academic regulatory institution for the Basque language. In their many efforts to research and support the language, they have conducted some invaluable work on dialects, including the \href{https://www.euskaltzaindia.eus/index.php?option=com_content&view=article&id=565&Itemid=466&lang=eu}{\textbf{Linguistic Atlas of Basque}}.

\begin{figure*}[!t]
  \centering
  \includegraphics[page=2, width=0.7\linewidth]{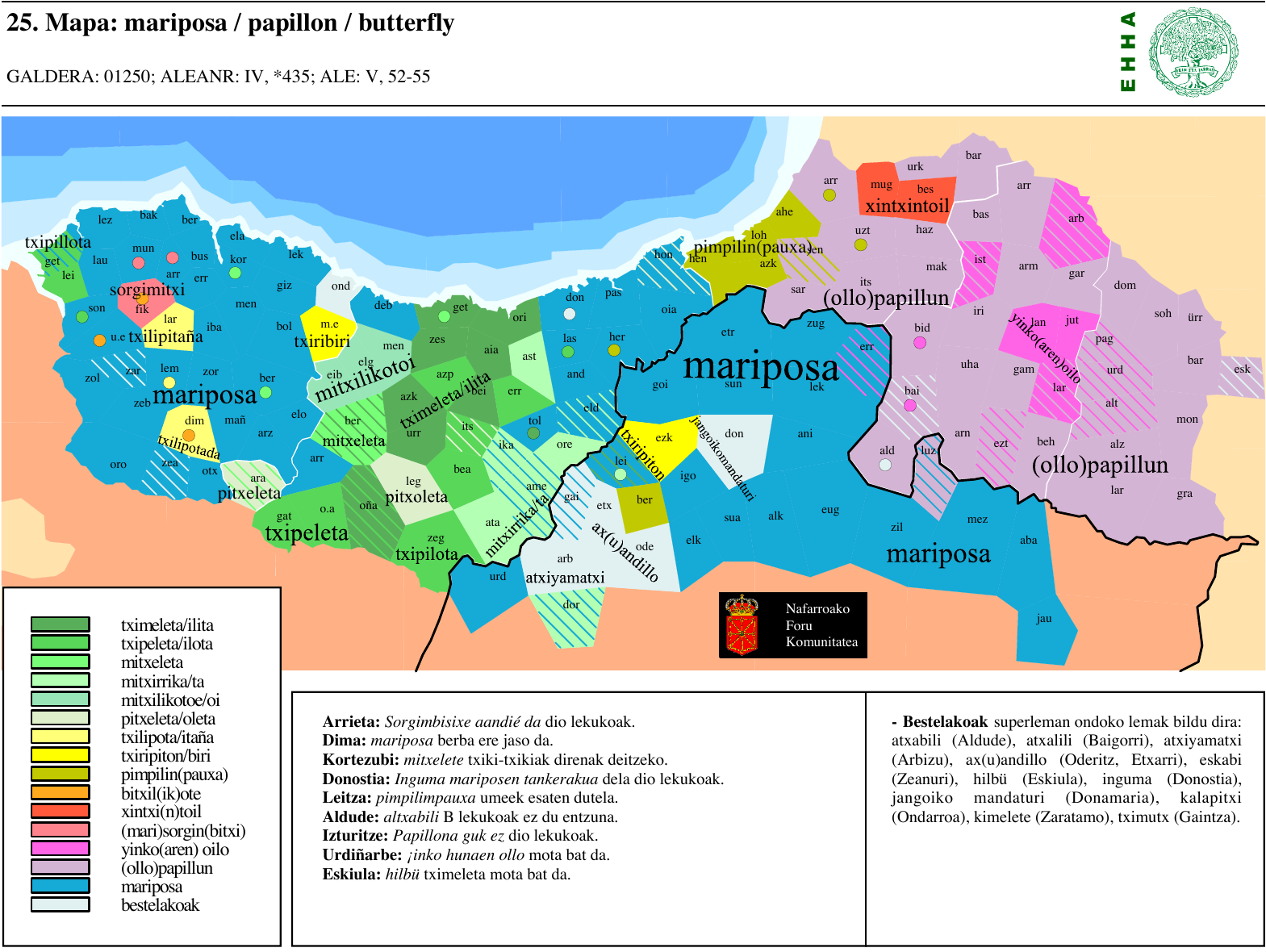}
  \caption{Example of a linguistic atlas.}
  \label{fig:tximeleta}
\end{figure*}

This atlas provides linguistic maps for many Basque words according to their geographical location, thus illustrating the big lexical variation that could be encountered through the several Basque-speaking regions. An example map is provided in Figure \ref{fig:tximeleta}.

\subsection{Dictionaries}
\label{sec:dict}


Some online dictionaries often include dialectal information when searching for specific words, such as Elhuyar or \href{https://www.euskaltzaindia.eus/index.php?option=com_hiztegianbilatu&task=hasiera&Itemid=1693&lang=eu}{\textbf{Euskaltzaindia}}.
The Elhuyar dictionary provides dialectal information about some words. The Euskaltzaindia dictionary provides not only geographical information for selected words, indicating their potential dialectal affiliation, but also enables searches to be filtered by dialect.

However, these standard dictionaries only provide information on dialectal words that are accepted in the standard, while other variations of dialectal words are absent. For instance, \textit{berba} (word) is accepted in the standard form and appears in dictionaries, while its possible variations, such as \textit{berbie} or \textit{berbi}, do not. 

Additionally, some dictionaries are designed to focus on a particular dialect, for example \href{https://www.labayru.eus/}{\textbf{LabayruHiztegia}}.
This is a Western-focused dictionary, with Standard-Western word pairs, as well as information on some multi-word expressions particular to this dialect.

\subsection{Grammar}
\label{sec:grammar}

Some efforts have been made to formalize the grammar rules from several Basque dialects. For example, the Labayru foundation has recently launched \href{https://gramatika.labayru.eus/}{\textbf{LabayruGramatika}},
a compilation of the grammatical rules from the Western dialect, with explanations as well as examples for every grammatical phenomenon. 

Furthermore, \citet{uria2012hizkeren} provided \href{https://ixa2.si.ehu.eus/atlas2/help.php?lang=eu}{\textbf{Basyque}}, an online resource to store, organize, manage and search for all the information concerning dialectal variation in, specifically, the North-Eastern Basque dialects, providing information that enables the syntactic analysis of these dialects.

\subsection{Dialectal Videos}
\label{sec:videos}

In an attempt to observe, preserve and analyze diachronic variation, some speakers have been interviewed, and the recordings have been uploaded to several web pages. These videos are annotated with the speakers' geographical origin. 

For instance, \href{https://ahotsak.eus/corpusa/}{\textbf{Ahotsak}}
provides thousands of videos of interviews with people from different generations, genders, backgrounds, and geographical locations. Some of these interviews have been transcribed and could act as a great resource for the analysis of oral Basque dialects. However, the videos with missing transcriptions still pose a great challenge.

Similar to Ahotsak, \href{https://www.mintzoak.eus/eu/}{\textbf{Mintzoak}}
compiles video-interviews of Northern Basque speakers, in order to keep the collective memory alive. Contrary to Ahotsak, this site does not contain transcriptions of the interviews. 

Additionally, \href{http://euskalkiak.eus/ikus_entzunezkoak.php}{\textbf{Euskalkiak.eus}}
also contains some videos based on the geographical location of the speakers, also with no transcriptions. 

\begin{figure*}[!t]
  \centering
  \includegraphics[width=\linewidth]{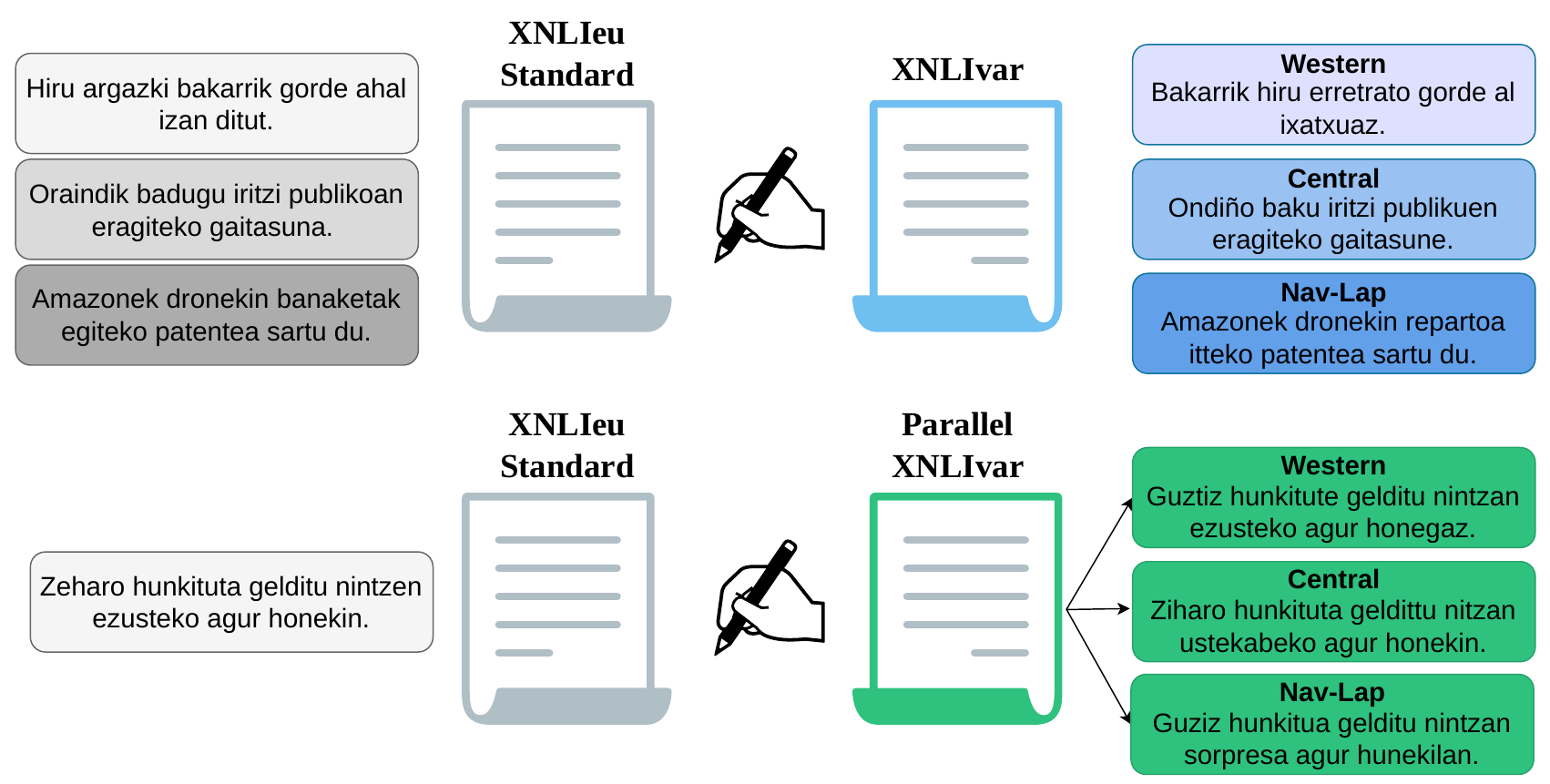}
  \caption{Illustration of XNLIeu dialectal adaptation. XNLIvar was a compilation of different instances in three different dialects. Parallel XNLIvar provides the same instances in three Basque dialects, offering completely parallel data.}
  \label{fig:xnli-example}
\end{figure*}

\section{Standard-to-Dialect Adapted Data}
\label{sec:adaptations}

In addition to these online dialectal resources, both manual and automatic adaptations from standard Basque into dialects have recently been developed. Standard-dialect parallel data constitute an important resource not only for NLP but also for studies in variationist linguistics and sociolinguistics.


\subsection{Manual Dialectal Adaptation}
\label{sec:manual-adaptation}

Manually adapting standard texts into dialects is to this day the most effective way to obtain gold standard data. However, despite its importance, relatively few studies have addressed this topic in Basque. This section describes the first manually adapted dataset into Basque dialects \citep{bengoetxea-etal-2025-lost}. Additionally, this paper presents a newly developed expansion of it, namely Parallel XNLIvar, also collected through manual adaptation. 

\begin{table}[!b]
    \centering
    \resizebox{\columnwidth}{!}{%
    \begin{tabular}{@{}l rl rl r@{}}
    \toprule
        \textbf{Dialect} & \textbf{Birthplace}  & \textbf{Age} & \textbf{Gender} & \textbf{Studies} \\ \midrule
         Western & Elorrio & 58 & Male & Translation\\
         Central & Arroa/Zumaia  & 34 & Female & Translation\\
         Nav-Lap & Donibane-Lohizune & 33 & Female & Translation \\
         \bottomrule

    \end{tabular}
    }
    \caption{Metadata from native Basque speakers who adapted the test partition of XNLIeu into dialects.}
    \label{tab:xnli-adaptors}
\end{table}

\paragraph{XNLIvar}

\citet{bengoetxea-etal-2025-lost} presented XNLIvar, a Natural Language Inference (NLI) dataset which contained data in three Basque dialects. This dataset was manually adapted from XNLIeu \citep{heredia-etal-2024-xnlieu}, an NLI dataset in Basque translated from the multilingual XNLI dataset \citep{conneau-etal-2018-xnli}. This dataset consists of Premise-Hypothesis pairs with entailment, neutral or contradiction relations. 

For the dialectal adaptation, \citet{bengoetxea-etal-2025-lost} adapted the native partition of XNLIeu, i.e., a 621-instance test set manually created in Basque. Although the dataset contained material from three Basque dialects (Western, Central, and Navarrese), it did not include fully parallel versions of the same content across dialects.

\paragraph{Parallel XNLIvar} 
This paper presents an expansion of XNLIvar by adapting the original XNLI test set of around 5000 instances into three different Basque dialects (Western, Central, and Navarrese-Lapurdian). This expansion was created in parallel, providing three fully dialectal versions of the XNLIeu test set, each corresponding to a different Basque dialect. This adaptation process is illustrated in Figure \ref{fig:xnli-example}.

The adaptation from standard to dialectal Basque has been manually carried out by native speakers of each dialect. Metadata from each native speaker is illustrated in Table \ref{tab:xnli-adaptors}, such as birthplace, age, gender and previous studies. 


All in all, this novel resource allows for a more thorough assessment of dialectal effects in NLI and facilitates per-dialect analysis through the inclusion of parallel data.

\begin{figure}[!t]
  \centering
  \includegraphics[width=\linewidth]{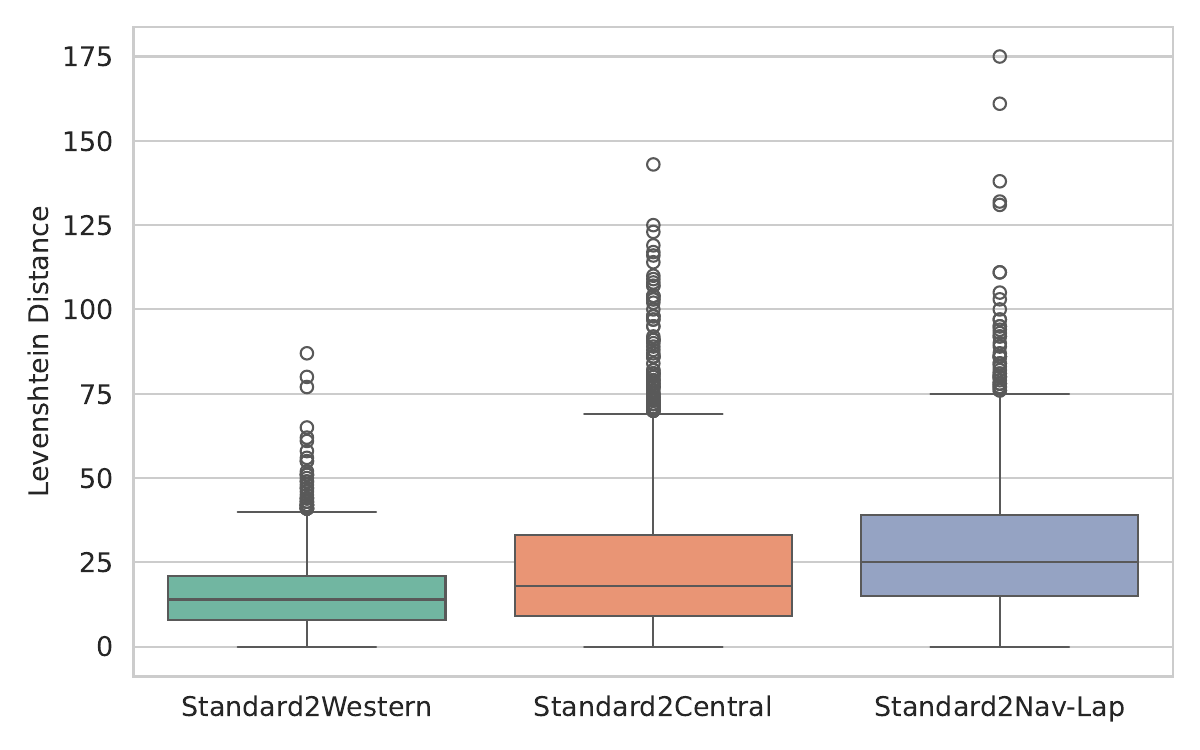}
  \caption{Levenshtein distance distribution for the three Parallel XNLIvar datasets.}
  \label{fig:distance}
\end{figure}

\subsubsection{Analysis of Parallel XNLIvar} 


Further analysis has been conducted to explore how different the dialectal datasets are from the standard. 
To do so, the Levenshtein distance between standard and dialectal sentences was calculated. This metric measures the minimum number of single-character insertions, deletions, or substitutions required to transform one string into another. The average distance results are presented in Table \ref{tab:distance}.

The results suggest that Navarrese-Lapurdian is the dialect most distant from the standard variety. This is consistent with Basque dialectological research, which has shown that peripheral dialects typically exhibit greater divergence \citep{michelena1981lengua}. 




Under this theoretical framework, the Western dialect should not be the closest to the standard variety, given that it is likewise considered a peripheral dialect. However, as we can see in Figure \ref{fig:distance}, the Central dialect seems to be further from the standard than the Western. This could be due to the distance distribution, as the Central dialect shows many sentences with big modifications, whereas the distances in the Western dataset remain consistently low across all sentences.



Further analysis of the two datasets shows that the linguist who did the Central adaptations frequently modified sentence word order. That could be why the Levenshtein distance for the Central dataset is higher than for the Western dataset, in which word order was largely preserved. This highlights the importance of robust metrics for measuring dialectal variation, as this distance metric estimates the transformation distance between sentences, but its effectiveness in capturing dialectalness seems to remain unclear, especially in free word-order languages like Basque.


\begin{table}[!b]
    \centering
    \begin{small}
    \begin{tabular}{lr}
    \toprule
        \textbf{Dialect} & \textbf{Distance} \\ \midrule
        Western &  15.56  \\
        Central & 24.73  \\
        Nav-Lap &  29.36 \\
        \bottomrule
    \end{tabular}
    \end{small}
    \caption{The average Levenshtein \textbf{Distance} of the Parallel XNLIvar dialectal datasets to the Standard.}
    \label{tab:distance}
\end{table}


\subsection{Automatic Dialectal Adaptation}
\label{sec:automatic-adaptation}

Given the high cost and time demands of manual dialectal adaptation, recent work has focused on automatically converting standard data into dialects, which is discussed in this section.

\paragraph{BasPhyCo}

\citet{bengoetxea2026physical} presented a physical commonsense reasoning dataset consisting of 356 instances of 5-sentence stories, which could be plausible or implausible. BasPhyCo was manually translated into standard Basque from its original Italian version \citep{pensa2024multi}. Additionally, the standard Basque dataset was automatically adapted into the Western dialect through few-shot prompting of Latxa-It-70B \citep{sainz-etal-2025-instructing}. 

Consequently, \citet{bengoetxea2026physical} provide two parallel versions of the same dataset: BasPhyCo and BasPhyCo\textsubscript{west}. The availability of standard–dialect parallel data allows to examine the impact of dialectal variation on physical commonsense reasoning.

\subsubsection{Evaluation of BasPhyCo\textsubscript{west}}

The automatic adaptation of BasPhyCo\textsubscript{west} was validated by a professional Basque linguist \citep{bengoetxea2026physical}. However, we sought to extend this evaluation to the adapted dataset, thereby measuring the actual dialectal value of the automatic adaptation for native Western dialect speakers. To do so, we have outlined the following manual evaluation framework.

\begin{figure*}[!t]
  \centering
  \includegraphics[width=\linewidth]{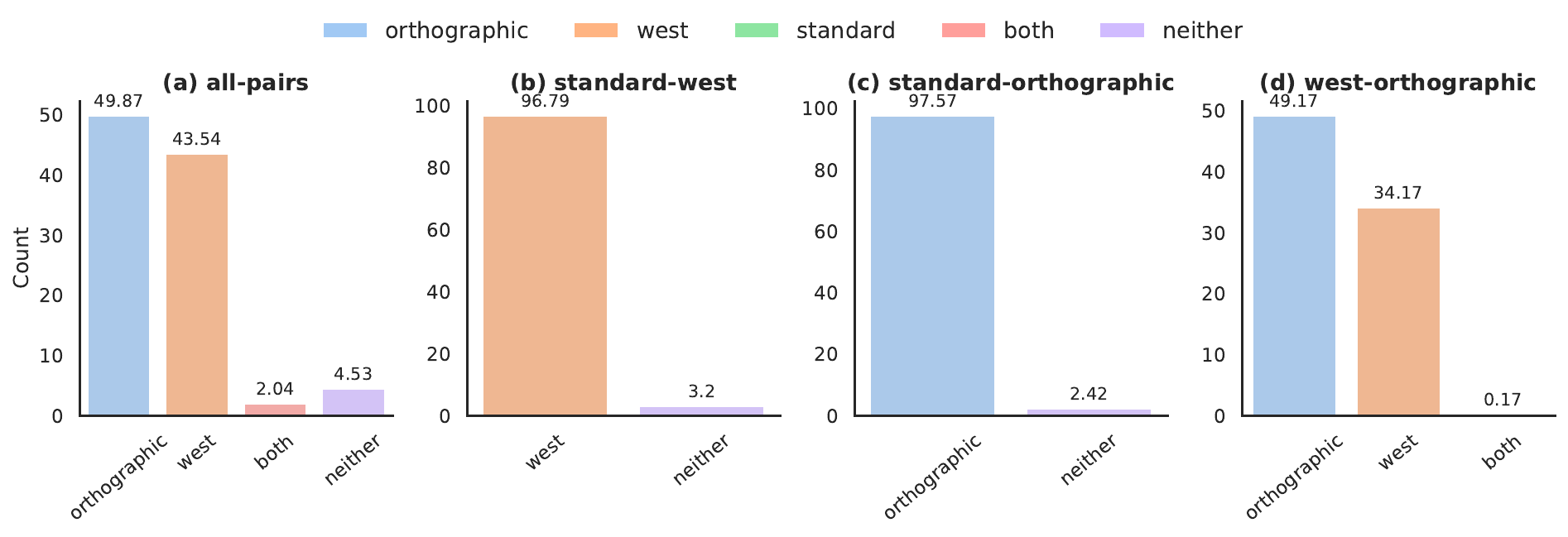}
  \caption{Manual evaluation results. From left to right, results for all sentence pairs (\textbf{all-pairs}), as well as results for different sentence pair combinations (\textbf{standard-west}, \textbf{standard-orthographic}, and \textbf{west-orthographic}).}
  \label{fig:results-manual-eval}
\end{figure*}

\paragraph{Evaluated datasets} 


The original prompt of \citet{bengoetxea2026physical} for the adaptation of BasPhyCo\textsubscript{west} explicitly allowed for non-standard orthographic modifications. We have adapted this prompt by eliminating the possibility of orthographic changes and obtained a new version, BasPhyCo\textsubscript{west-new}. Both prompts used for the adaptations are provided in the Appendix.

Therefore, our evaluation was done on three versions of the dataset: the \textbf{Standard} (BasPhyCo), the Western \textbf{Orthographic} (BasPhyCo\textsubscript{west}), and a novel \textbf{Western}, non-standard orthographic version (BasPhyCo\textsubscript{west-new}). The following examples illustrate two dialectal sentences, both with standard and non-standard orthography.

\begin{description}
    \item[Standard] Teknikaria ez da oraindi\textbf{n}o etorri
    \item[Non-Standard] Teknikarixa ez dau oraindi\textbf{ñ}o etorri
\end{description}

We can observe that the non-standard version includes the letter \textit{ñ} in \textit{oraindiño}, while the standard version follows standard orthographic rules and favors the use of the letter \textit{n}.



\paragraph{Description of the task} 

The evaluation was proposed as a pairwise comparison task. The pairs of sentences to be evaluated were constructed as a Cartesian product, i.e., every sentence from dataset A was paired with every sentence in dataset B. Thus, the combination of datasets A $\times$ B is the following:

\begin{equation}
    {\displaystyle A\times B=\{(a,b)\mid a\in A\ {\mbox{ and }}\ b\in B\}.}
\end{equation}

As we have three different datasets (A, B and C), the Cartesian product for all three dataset combinations was conducted and subsequently concatenated. Given that $\times$ is the Cartesian product, the pairs to be evaluated were constructed as: 

\begin{equation}
{\displaystyle (A\times B)\cup(A\times C)\cup(B\times C)}
\end{equation}

For the manual evaluation, 180 sentence pairs from each dataset combination were randomly sampled, obtaining a total of 540 sentence pairs for evaluation. 


The task consisted of annotating which sentence from each pair was closer to the Western dialect of Basque. The evaluators were given four options: \textit{sentence A} is more Western, \textit{sentence B} is more Western, they are \textit{both} Western, or \textit{neither} is Western. 

\paragraph{Annotators}

The evaluation was made by two different native Western speakers, with the same educational background, but different age, gender, and birthplace. The metadata for each evaluator is presented in Table \ref{tab:manual-eval}. 

The Inter-Annotator Agreement (IAA) between the two evaluators has been computed, with a Cohen's Kappa of 0.71, which constitutes a substantial agreement score. 

\begin{table}[!t]
    \centering
    \resizebox{\columnwidth}{!}{%
        \begin{tabular}{@{}lrl rl r@{}}
    \toprule
        \textbf{Dialect} & \textbf{Birthplace}  & \textbf{Age} & \textbf{Gender} & \textbf{Studies} \\ \midrule
         Western & Oñati & 33 & Female & Translation \\
         Western &  Elorrio & 59 & Male & Translation \\
         \bottomrule

    \end{tabular}
    }
    \caption{Metadata from the native Western Basque speakers for the manual evaluation of the automatic adaptation.}
    \label{tab:manual-eval}
\end{table}

\paragraph{Quantitative results} 

The general results, as well as the results for different sentence-pair types, are illustrated in Figure \ref{fig:results-manual-eval}. 

The results for all sentence types (\ref{fig:results-manual-eval}a) show that the orthographic dataset has the most Western features, although it is closely followed by the Western dataset. No Standard sentence was marked as dialectal, which consolidates the confidence of the evaluation quality. Evaluators marked some doubtful sentence pairs as \textit{Neither} or \textit{Both}, which are examined in the following analysis section. 

Regarding the results for the different sentence-pair types, we can observe that when the dialectal sentences were paired with a standard sentence (i.e standard-west and standard-orthographic sentence pairs), the evaluators always chose the adapted sentences over the standard (Figure \ref{fig:results-manual-eval}b and \ref{fig:results-manual-eval}c). Additionally, when having to choose between the two adapted sentence types (west-orthographic), the evaluators deemed the Orthographic dataset more dialectal, but still closely followed by Western sentences (\ref{fig:results-manual-eval}d).

Consequently, this manual evaluation has shown that both automatically adapted datasets (Orthographic and Western) contain considerable dialectal features compared to the standard version. Additionally, explicitly stating in the adaptation prompt that non-standard orthographic changes are possible seems to generate even more dialectal adaptations, according to this evaluation. 


   
     



\paragraph{Analysis}

During the pairwise comparison, evaluators were also given two extra options: \textit{Both} and \textit{Neither}. It can be observed in Figure \ref{fig:results-manual-eval}b and \ref{fig:results-manual-eval}c that all \textit{Neither} labeled pairs seem to occur when one of the sentences in the pair is from the standard dataset. This reveals that the automatic adaptation sometimes failed to transform the standard sentences into their dialectal form. For example, the following sentences from the adapted datasets do not contain dialectal features: 

\begin{description}
    \item[Ortho] Koldo esnatu da. (\textit{Koldo has woken up})
    \item[West] Izotz-ontziak urtu dira. (\textit{The ice-cubes have melted})
\end{description}

Similarly, all \textit{Both} instances seem to occur when comparing the two automatically adapted datasets. This highlights that the Standard dataset is not biased towards dialectal language, as evaluators have not once considered a standard sentence to be dialectal.  

Furthermore, disagreements between annotators have also been examined. The majority of these instances occur on difficult sentence pairs, i.e., sentences that contain little to no dialectal features. Further analysis of these instances has revealed some slight annotator bias, as one annotator considered \textit{heldu} (arrive) a western marker, while the other did not. 

\begin{description}
    \item[Ortho] Ane berandu \textbf{helduko} da etxera. (\textit{Ane arrived late home})
    \item[Standard] Ane berandu \textbf{iritsiko} da etxera. (\textit{Ane arrived late home})
\end{description}

Finally, although error identification was not an annotation requirement, evaluators noted several errors in the dialectal adaptations. This points to clear limitations in the current dialectal adaptation capabilities of LLMs, which require further investigation.

\section{Conclusion}

This work presents a comprehensive collection of Basque dialectal data, categorized into two groups. First, online dialectal data and resources have been presented, grouped according to their domain, such as news, legal documents, informal tweets, dictionaries, grammar collections or even audiovisual resources. Secondly, standard datasets that were adapted into dialects have been described, both manually and automatically.

This paper has introduced two main contributions: (i) Parallel XNLIvar, a novel manually adapted parallel dialectal dataset for NLI. (ii) A manual evaluation of the automatically adapted BasPhyCo\textsubscript{west} dataset, outlined as a pairwise comparison task and performed by native dialectal speakers.


\section*{Limitations}

The main limitation regarding online sources is that some data cannot be used due to licensing issues. 

Additionally, automatic dialectal adaptation has been evaluated for only one Basque dialect, while other dialects remain unexamined. 

Finally, the manual evaluation of the automatic dialectal adaptation has revealed some errors, which presents an evident limitation in the possibility of silver data creation through automatic adaptation.


\section*{Acknowledgments}
This work has been supported by the HiTZ center and the Basque Government (Research group funding IT-1805-22).
Jaione Bengoetxea is funded by the Basque Government pre-doctoral grant (PRE\_2024\_1\_0028).

We also acknowledge the following MCIN/AEI/10.13039/501100011033 project: (i) DeepMinor (CNS2023-144375) and European Union NextGenerationEU/PRTR and (ii) DeepThought (PID2024-159202OB-C21) funded by ERDF, EU.

\section*{Bibliographical References}\label{sec:reference}

\bibliographystyle{lrec2026-natbib}
\bibliography{lrec2026-example}

\appendix

\begin{figure*}[!b]
  \centering
  \includegraphics[width=1.03\linewidth]{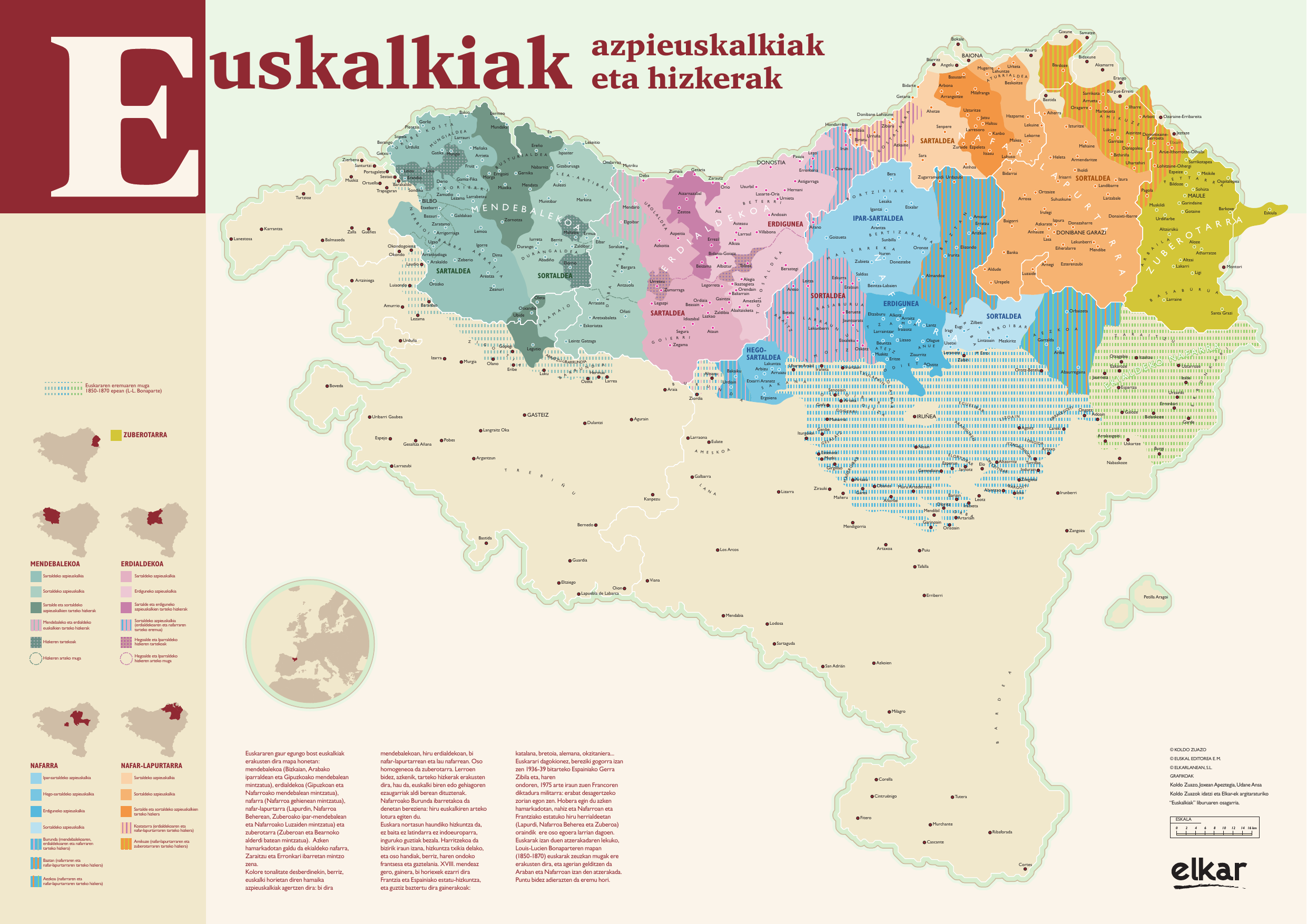}
  \caption{The map of the classification of Basque dialects according to \citet{zuazu2008}.}
  \label{fig:euskalkiak-map}
\end{figure*}

\newpage
\section{A Map of Basque Dialects}

The map in Figure \ref{fig:euskalkiak-map} shows the Basque dialect classification according to \citet{zuazu2008}.

\section{Dialectal Adaptation Prompts}

The prompts used to prompt Latxa-3.1-70B-It for the dialectal adaptation are provided below, both for the non-standard orthographic changes (Figure \ref{fig:prompt}) and for the standard orthographic changes (Figure \ref{fig:prompt-2}).

\begin{small}
    
\begin{figure*}[!b]
\centering
\begin{tcolorbox}[promptbox]

I will give you three versions of a story. Each version has five sentences. Some sentences are identical across versions. You need to adapt this text so that it includes Bizkaian dialectal features. \textbf{You can use non-standard orthography. Try to make it as similar as possible to oral language.} \\

\textbf{Task:}

\begin{enumerate}
    \item First, list all unique sentences across all three stories.
    \item Adapt each unique sentence exactly once into the Bizkaian dialect.
    \item Then reconstruct the three stories with the translations, making sure that any identical source sentence always has the identical translation.
    \item If there are more than three stories, repeat the same process for all of them.
\end{enumerate}

\textbf{Format:} \\

This is an example of an standard (INPUT) instance and an example of the dialectal (OUTPUT) adaptation that you need to do: \\

\textit{Standard:} \\

\hspace{0.3cm}\textit{STORY1:} ['Mikel lanera joan da', 'Mikelek ordenagailua piztu du', 'Mikelek mezuak irakurri ditu', 'Mikelek mezuak erantzun ditu', 'Mikel etxera joan da']

\hspace{0.3cm}\textit{STORY2:} ['Mikel lanera joan da', 'Mikelek mezuak erantzun ditu', 'Mikelek mezuak irakurri ditu', 'Mikelek ordenagailua piztu du', 'Mikel etxera joan da']

\hspace{0.3cm}\textit{STORY3:} ['Mikel lanera joan da', 'Mikelek ordenagailua itzali du', 'Mikelek mezuak irakurri ditu', 'Mikelek mezuak erantzun ditu', 'Mikel etxera joan da'] \\

\textit{Dialectal:} \\

\hspace{0.3cm}\textit{STORY1:} ['Mikel lanera jun de', 'Mikelek ordenagaillua piztu dau', 'Mikelek mesuek irakurri dauz', 'Mikelek mesuek erantzun dauz', 'Mikel etxera jun de']

\hspace{0.3cm}\textit{STORY2:} ['Mikel lanera jun de', 'Mikelek mesuek erantzun ditu', 'Mikelek mesuek irakurri dauz', 'Mikelek ordenagaillua piztu dau', 'Mikel etxera jun de']

\hspace{0.3cm}\textit{STORY3:} ['Mikel lanera jun de', 'Mikelek ordenagaillua amatatu dau', 'Mikelek mesuek irakurri dauz', 'Mikelek mesuek erantzun dauz', 'Mikel etxera jun de'] \\

Output only the reconstructed stories in the exact same format as the input. Do not output explanations, steps, or commentary.
\end{tcolorbox}
\caption{Dialectal adaptation prompt for the non-orthographic dataset version (BasPhyCo\textsubscript{west}).}
\label{fig:prompt}
\end{figure*}
\end{small}

\begin{small}
    
\begin{figure*}[!b]
\centering
\begin{tcolorbox}[promptbox]

I will give you three versions of a story. Each version has five sentences. Some sentences are identical across versions. \textbf{You need to adapt this text so that it includes Bizkaian dialectal features.}\\

\textbf{Task:}

\begin{enumerate}
    \item First, list all unique sentences across all three stories.
    \item Adapt each unique sentence exactly once into the Bizkaian dialect.
    \item Then reconstruct the three stories with the translations, making sure that any identical source sentence always has the identical translation.
    \item If there are more than three stories, repeat the same process for all of them.
\end{enumerate}

\textbf{Format:} \\

This is an example of an standard (INPUT) instance and an example of the dialectal (OUTPUT) adaptation that you need to do: \\

\textit{Standard:} \\

\hspace{0.3cm}\textit{STORY1:} ['Mikel lanera joan da', 'Mikelek ordenagailua piztu du', 'Mikelek mezuak irakurri ditu', 'Mikelek mezuak erantzun ditu', 'Mikel etxera joan da']

\hspace{0.3cm}\textit{STORY2:} ['Mikel lanera joan da', 'Mikelek mezuak erantzun ditu', 'Mikelek mezuak irakurri ditu', 'Mikelek ordenagailua piztu du', 'Mikel etxera joan da']

\hspace{0.3cm}\textit{STORY3:} ['Mikel lanera joan da', 'Mikelek ordenagailua itzali du', 'Mikelek mezuak irakurri ditu', 'Mikelek mezuak erantzun ditu', 'Mikel etxera joan da'] \\

\textit{Dialectal:} \\

\hspace{0.3cm}\textit{STORY1:} ['Mikel lanera jun de', 'Mikelek ordenagaillua piztu dau', 'Mikelek mesuek irakurri dauz', 'Mikelek mesuek erantzun dauz', 'Mikel etxera jun de']

\hspace{0.3cm}\textit{STORY2:} ['Mikel lanera jun de', 'Mikelek mesuek erantzun ditu', 'Mikelek mesuek irakurri dauz', 'Mikelek ordenagaillua piztu dau', 'Mikel etxera jun de']

\hspace{0.3cm}\textit{STORY3:} ['Mikel lanera jun de', 'Mikelek ordenagaillua amatatu dau', 'Mikelek mesuek irakurri dauz', 'Mikelek mesuek erantzun dauz', 'Mikel etxera jun de'] \\

Output only the reconstructed stories in the exact same format as the input. Do not output explanations, steps, or commentary.

\end{tcolorbox}
\caption{Dialectal adaptation prompt for the orthographic dataset version (BasPhyCo\textsubscript{west-new}).}
\label{fig:prompt-2}
\end{figure*}
\end{small}


\end{document}